\RequirePackage{fix-cm}
\documentclass[twocolumn]{svjour3}          
\smartqed  
\usepackage{graphicx}
\usepackage{times}
\usepackage{epsfig}
\usepackage{epstopdf}
\usepackage{amsmath}
\usepackage{amssymb}
\usepackage[labelfont=bf]{caption}
\usepackage{tabularx}
\usepackage[table]{xcolor}
\usepackage[noend]{algpseudocode}

\usepackage{booktabs}
\usepackage{makecell}
\usepackage{multirow}
\usepackage{mmstyles}
\usepackage{siunitx}
\usepackage{marvosym}
\usepackage{natbib}
\definecolor{citecolor}{HTML}{0071bc}
\definecolor{tabhighlight}{HTML}{e5e5e5}
\usepackage[colorlinks,citecolor=citecolor]{hyperref}
\usepackage{bbm}
\usepackage{pifont}
\usepackage{subcaption}
\usepackage{float}

\definecolor{ForestGreen}{rgb}{0.13, 0.55, 0.13}
\definecolor{Green}{rgb}{0.0, 0.5, 0.0}
\definecolor{green(munsell)}{rgb}{0.0, 0.66, 0.47}
\definecolor{green(ryb)}{rgb}{0.4, 0.69, 0.2}
\definecolor{green(pigment)}{rgb}{0.0, 0.65, 0.31}
\definecolor{GrayXMark}{gray}{0.6}

\usepackage[ruled]{algorithm2e}
\newcommand{\f}[1]{\textbf{#1}} 
\newcommand{\s}[1]{\underline{#1}} 
\newenvironment{smalltable}[1]{ \footnotesize \begin{tabular}{#1} }{ \end{tabular} }

\usepackage{pifont}
\newcommand{\cmark}{\ding{51}}
\newcommand{\xmark}{\ding{55}}
\def\aps{$\text{AP}_\text{S}$}
\def\apm{$\text{AP}_\text{M}$}
\def\apl{$\text{AP}_\text{L}$}

\makeatletter
\def\BState{\State\hskip-\ALG@thistlm}
\makeatother
\definecolor{myGreen}{HTML}{33FF00}
\definecolor{myRed}{HTML}{FF3030}
\definecolor{myGrey}{HTML}{AA5555}
\definecolor{myWhite}{HTML}{FFFFFF}
\definecolor{maroon}{cmyk}{0,0.87,0.68,0.32}
\definecolor{petr}{HTML}{5555FF}
\definecolor{josef}{HTML}{FF3030}

\journalname{International Journal of Computer Vision}

\makeatletter
\def\makeheadbox{%
  \hbox to0pt{\vbox{\baselineskip=10dd\hrule\hbox
    to\hsize{\vrule\kern3pt\vbox{\kern3pt
    \hbox{\bfseries\@journalname}
    \kern3pt}\hfil\kern3pt\vrule}\hrule}%
  \hss}}
\makeatother

\begin{document}
\begin{sloppypar}

\title{EdgeSAM: Prompt-In-the-Loop Distillation for SAM}


\author{Chong Zhou          \and
        Xiangtai Li         \and
        Chen Change Loy *   \and
        Bo Dai
}


\institute{Chong Zhou \at
              S-Lab, Nanyang Technological University, Singapore \\
              \email{chong003@ntu.edu.sg}
           \and
           Xiangtai Li \at
              S-Lab, Nanyang Technological University, Singapore \\
              \email{xiangtai.li@ntu.edu.sg}
           \and
           Chen Change Loy \at
              Corresponding Author \\
              S-Lab, Nanyang Technological University, Singapore \\
              \email{ccloy@ntu.edu.sg}
           \and
           Bo Dai \at
              The University of Hong Kong, China \\
              \email{bdai@hku.hk}
}

\date{Received: 16 January 2025 / Accepted: 4 August 2025}

\maketitle

\begin{abstract}
    This paper presents EdgeSAM, an accelerated variant of the Segment Anything Model (SAM), optimized for efficient execution on edge devices with minimal compromise in performance. 
    Our approach involves distilling the original ViT-based SAM image encoder into a purely CNN-based architecture, better suited for edge devices. 
    We carefully benchmark various distillation strategies and demonstrate that task-agnostic encoder distillation fails to capture the full knowledge embodied in SAM. 
    To overcome this bottleneck, we include both the prompt encoder and mask decoder in the distillation process, with box and point prompts in the loop, so that the distilled model can accurately capture the intricate dynamics between user input and mask generation. 
    To mitigate dataset bias issues stemming from point prompt distillation, we incorporate a lightweight module within the encoder.  
    As a result, EdgeSAM achieves a 37-fold speed increase compared to the original SAM, and it also outperforms MobileSAM/EfficientSAM, being over 7 times as fast when deployed on edge devices while enhancing the mIoUs on COCO and LVIS by 2.3/1.5 and 3.1/1.6, respectively. It is also the first SAM variant that can run at over 30 FPS on an iPhone 14. 
    Code and demo are available \href{https://www.mmlab-ntu.com/project/edgesam}{here}.
\end{abstract}

\section{Introduction}
\label{sec:intro}

In this study, we explore the feasibility of deploying the Segment Anything Model (SAM)~\citep{sam} directly on edge devices, such as smartphones, to enable real-time interactive segmentation and facilitate its integration in various downstream tasks. Addressing this challenge is non-trivial and non-achievable with the usual engineering efforts of compressing the model. First, SAM was not designed for on-device deployment. It has a total parameter count of 641M and requires 2735 GFLOPs under its fixed $1024 \times 1024$ input resolution. We failed to run SAM on an iPhone 14 due to its large computation and memory consumption. Even on an NVIDIA 2080 Ti, the throughput is just four images per second.

A seemingly straightforward solution to speed up is to replace SAM's huge ViT-based image encoder with a more compact version, as demonstrated by MobileSAM~\citep{mobile_sam}. This approach indeed leads to a considerable speed boost, approximately 23 times faster, but it also substantially reduces the performance. For instance, the mask mIoU on the COCO dataset~\citep{coco} drops from 77.3 to 74.4 with ground-truth boxes as the box prompts. Moreover, when deployed on edge devices, the speed of MobileSAM is still far from being real-time. For example, on iPhone 14, its throughput is only five images per second. A more recent work, EfficientSAM~\citep{efficient_sam} follows the same architecture as MobileSAM but further leverages the masked image pre-training to boost the performance. However, it runs at the same speed as MobileSAM and consumes huge computational costs during training.

In our extensive empirical investigations, it became unexpectedly clear that the key factor in knowledge distillation from SAM is not the incorporation of losses specifically tailored for dense tasks~\citep{liu2019structured,shu2021channel} or query-based detectors~\citep{huang2022teachdetr,chen2022d,chang2023detrdistill}. Rather, it is the strategic selection of prompts during the distillation process that holds paramount importance. Addressing this, we introduce a novel approach: a dynamic \textbf{prompt-in-the-loop} strategy. This technique effectively enables the student model to assimilate the intricate knowledge encapsulated within SAM. It involves actively aligning the student model with the multi-grained output masks of SAM and iteratively introducing new prompts in regions where the student model exhibits inaccuracies. These tailored prompts guide the mask decoder, emphasizing areas of incorrect segmentation, thereby enhancing the learning process. Furthermore, we investigated various training configurations, examining different types of prompts, the impacts of freezing encoder/decoder components, the selection of distillation targets, \etc. The specifics and findings of these explorations are detailed in the experimental section.

Additionally, we conducted thorough ablation studies focusing on the selection of the backbone architecture, particularly in view of the throughput-performance balance crucial for on-device deployment. We find that purely CNN-based architectures emerge as the more advantageous choice to ViT-based backbones for achieving the optimal trade-off. This is attributed to the current landscape of on-device AI accelerators, such as Apple Neural Engine (ANE), which are predominantly optimized for CNNs rather than ViT architectures. This observation also underscores the versatility of our proposed prompt-aware knowledge distillation approach, highlighting its applicability across diverse architectures.

In our final observations, we note that SAM, having been trained on a dataset with multi-grained annotations, encounters challenges in resolving the granularity of output when faced with ambiguous prompts, such as a single point. This is particularly evident when SAM is prompted with center points on the COCO dataset during evaluations; the model does not consistently produce instance-level masks, but rather part-level masks. This issue becomes more pronounced when SAM functions as the teacher model. To address this, we propose a simple yet effective module designed to explicitly discern and adapt to the granularity priors specific to a given test set or application scenario. This module enhances SAM's ability to interpret and respond to varying levels of prompt ambiguity accurately.

Consequently, our EdgeSAM model achieves a remarkable performance boost, operating \emph{37 times} faster than the original SAM and \emph{0.6 times} more swiftly than MobileSAM and EfficientSAM on an NVIDIA 2080 Ti GPU. Notably, on the iPhone 14, EdgeSAM leverages its purely CNN-based backbone~\citep{rep_vit} to process image encoding in a mere 26 ms per image. This rate is \emph{over 7 times faster} compared to the performance of MobileSAM and EfficientSAM on the same platform, marking it as the first SAM variant capable of real-time operation on edge devices. In terms of accuracy, EdgeSAM closely parallels the original SAM in box-prompt performance across the SA-1B~\citep{sam}, COCO~\citep{coco}, and LVIS~\citep{lvis} datasets and surpasses MobileSAM and EfficientSAM by an average of 1-2 mIoU across these datasets. This capability to run SAM in real-time on edge devices opens up a multitude of possibilities for downstream applications, including on-device video editing and video instance segmentation.

\section{Related Work}
\label{sec:related_work}

\noindent
\textbf{Efficient Model Design.} This research direction mainly focuses on designing efficient CNNs~\citep{mnetv1,mnetv2,mnetv3,squeezenet,shufflenetv1,shufflenetv2,ghostnet}, transformers~\citep{edgenext,edgevit,mvitv1}, and their mixed architecture~\citep{EMOiccv23,mocovit,mobileformer,nextvit}, with the goal of visual representation learning. 
Our work also adopts efficient models as the image encoder but is orthogonal to these works as it can be applied to various efficient backbones.

\begin{figure*}
    \centering
    \includegraphics[width=0.8\textwidth]{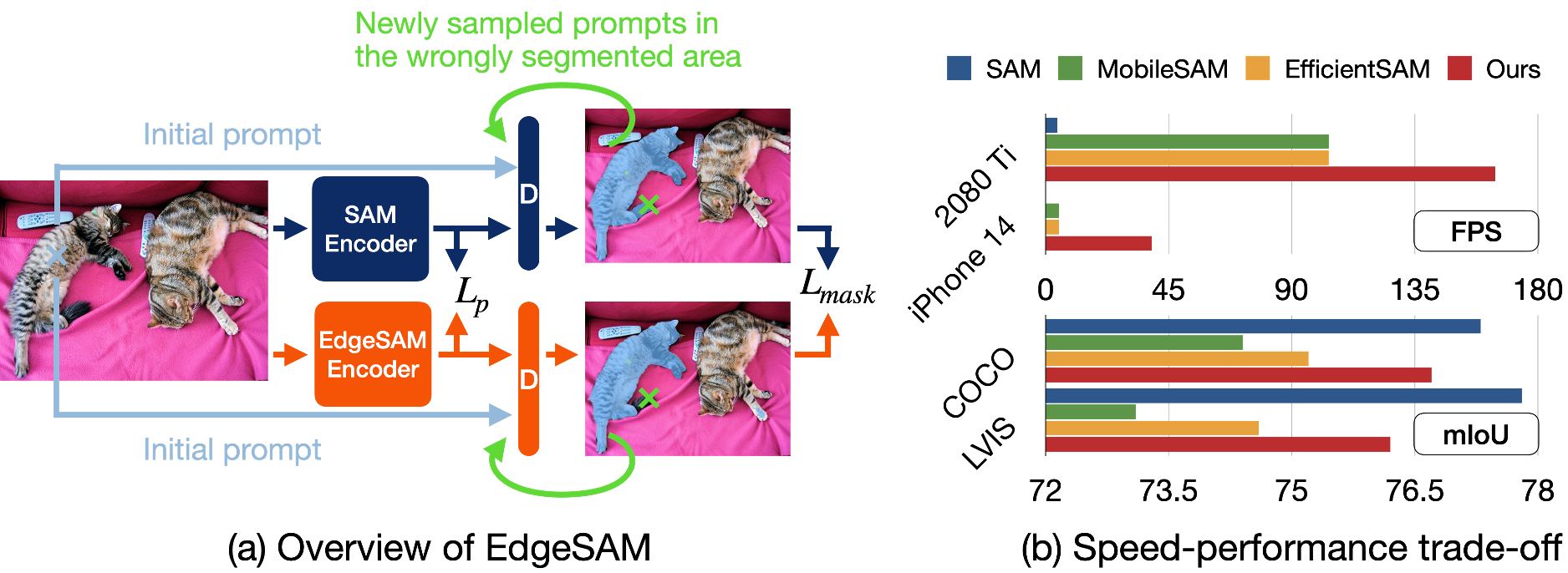}
    \caption{Figure (a) shows the overview of EdgeSAM. We first apply encoder-only knowledge distillation between the output feature of the image encoder of SAM and EdgeSAM. Then, we adopt prompt-in-the-loop knowledge distillation, which interactively samples point prompts from the wrongly segmented areas. Note that the initial prompt can also be in a box. In (b), we show its throughput and the mIoU performance on the COCO and LVIS datasets with box prompts compared with SAM, MobileSAM, and EfficientSAM.}
    \label{fig:teaser}
\end{figure*}

\noindent
\textbf{Knowledge Distillation in Detection and Segmentation.} 
The majority of research in knowledge distillation has been centered on classification tasks~\citep{hinton2015distilling,guan2020differentiable,xie2020self,yuan2020revisiting,zhang2020distilling}. Several studies~\citep{xie2018improving,liu2019structured,shu2021channel,wang2020intra,zhang2020improve,wang2019distilling} have applied knowledge distillation techniques to dense prediction tasks like semantic segmentation and object detection. Common approaches in these works involve leveraging pixel-wise correlations or channel-wise interactions between dense features of teacher and student models. Recently, there has been an increased interest in developing specialized knowledge distillation losses for query-based detectors such as DETR~\citep{detr}, as evidenced by works like~\citep{huang2022teachdetr,chen2022d,chang2023detrdistill}. MobileSAM~\citep{mobile_sam}, closely related to our work, implements pixel-wise feature distillation between the SAM encoder and a compact backbone and works following this direction include EfficientViT-SAM~\citep{efficientvit_sam}, RepViT-SAM~\citep{repvit-sam}, and NanoSAM~\citep{nanosam}. However, they do not address the prompt encoder and mask decoder, leading to a significant performance discrepancy compared to the original SAM. In particular, RepViT-SAM follows MobileSAM and replaces its image encoder with RepViT, thus yielding the same speed as EdgeSAM. However, because it only adopts the encoder-only KD, its performance falls behind our work. FastSAM~\citep{fast_sam}, on the other hand, trains a YOLACT-based~\citep{yolact} instance segmentation model using the SA-1B dataset and employs heuristic rules for post-process object selection, a method that aligns marginally with the SAM principles. 
EfficientSAM~\citep{efficient_sam} has achieved a great speed-performance trade-off through masked image pre-training, but it consumes a huge computational cost during training, and as it uses the same image encoder as MobileSAM, it runs no faster than MobileSAM. Our work further explores the setting where both training and inference budgets are more limited. 

Recently, SAM 2~\citep{sam2} extends SAM to the video domain, and several works optimize SAM 2 for efficient inference~\citep{efficienttam, edgetam}. Our work can also contribute to the SAM pre-training stage for these methods. The design philosophy of the proposed prompt-in-the-loop knowledge distillation is also related to refinement-based dense prediction methods such as Cascade-RCNN~\cite{cascade_rcnn} and RefineMask~\cite{refinemask}, but with a focus on distillation instead of architecture.

\noindent
\textbf{Efficient Segmentation Models.} Prior studies in efficient segmentation~\citep{ICnet,SFnet,bisenet,hu2023you,hong2021lpsnet,wan2023seaformer,yu2021bisenetv2,Li2022SFNetFA,espnetv2,mehta2018espnet,hong2021deep} have predominantly concentrated on close-set segmentation within specific domains, with a significant portion of this research~\citep{SFnet,hu2023you,espnetv2} specifically targeting driving scenarios. More recently, a few works~\citep{zhang2022topformer, wan2023seaformer} have ventured into designing segmentation models suitable for on-device implementation, capable of running efficiently on mobile platforms. However, the realm of on-device interactive segmentation remains largely unexplored. MobileSAM~\citep{mobile_sam} represents an initial attempt, but it still encounters challenges regarding computational efficiency and a noticeable decline in performance, indicating a substantial opportunity for further advancements in this direction. More related research can be found in this survey~\citep{efficient_sam_survey}.

\section{Method}
\label{sec:method}
In this section, we first briefly introduce SAM (Sec.~\ref{sec:sam}), followed by our proposed method, EdgeSAM, that slims SAM by encoder distillation (Sec.~\ref{sec:encoder-kd}), prompt-in-the-loop distillation (Sec.~\ref{sec:prompt-kd}), and a lightweight module that embeds the granularity preferences (Sec.~\ref{sec:rpn}).

\subsection{Preliminaries: SAM}
\label{sec:sam}
The SAM model consists of three components: image encoder, prompt encoder, and mask decoder. In particular, the image encoder takes up the most computational cost and parameters. It follows the backbone design of ViTDet~\citep{vitdet}, which essentially is a hierarchical vision transformer. The input and output sizes of the image encoder are kept fixed at $(3,1024,1024)$ and $(256,64,64)$.

SAM can handle four types of prompts, including point, box, mask, and text. Points and boxes are considered as the sparse prompts, which are encoded with random positional embeddings~\citep{pos_emb} summed with a type indicator token. Masks and free-form texts are embedded with a tiny CNN and the text encoder of CLIP~\citep{clip}, respectively. In this paper, we focus on the point and box prompts. Note that, when no prompt is given to SAM, SAM can operate in the segment everything mode, where it generates a grid of positive points and regards each point as a prompt.

The mask decoder takes as input the image feature, embedded point and/or box prompts, mask prediction tokens, and an IoU prediction token. All the inputs will be mixed and encoded with a lightweight two-way transformer. As a result, each mask token is transformed into a dynamic linear classifier, which can compute the foreground probability of each spatial location of the image feature. The IoU token yields the mask confidence score for each mask.

Apart from the interactive prompt-based model design, the SA-1B dataset also contributes significantly to SAM's zero-shot generality. It is the largest segmentation dataset to date, with over 1 billion mask annotations across 11 million images. Here, we highlight several characteristics of SA-1B to better understand SAM's capabilities: (1) the mask annotations are obtained by prompting an ambiguity-aware model, that is trained on semi-automatically labeled data, with grid points; (2) the masks are class-agnostic; (3) and the annotations are multi-grained in both instance and part levels. As we will discuss later, those properties of SA-1B make distillation on SAM different and more challenging than previous segmentation models.

\subsection{EdgeSAM}

The goal of EdgeSAM is to transfer the capabilities of SAM into a much more compact model, which makes deployment on edge devices feasible. In particular, as shown in Fig.~\ref{fig:teaser}, EdgeSAM maintains the encoder-decoder architecture of SAM and aims to preserve the performance of zero-shot interactive segmentation with box and point prompts. We train EdgeSAM with only 1\% data from the SA-1B dataset and evaluate its zero-shot transferability on the COCO~\citep{coco} and LVIS~\citep{lvis} datasets.

\subsubsection{Encoder-Only Knowledge Distillation}
\label{sec:encoder-kd}
Inspired by MobileSAM~\citep{mobile_sam}, we adopt a pixel-wise feature distillation loss $L_p$ between the image encoder $T_{enc}$ of SAM and an efficient network $S_{enc}$ as follows:
\begin{align}
    L_p = \text{MSE}(T_{enc}(I), S_{enc}(I)),
\end{align}
where $I$ denotes the input image. Since the downsampling stride and feature channels of the student model and SAM image encoder are not aligned, MobileSAM removes the downsampling operations in the last two stages of the student model and uses a projection layer to align the channel dimension. We also use the projection layer for channel alignment, but we keep the downsampling layers unchanged. Instead, we construct a tiny FPN~\citep{fpn} that upsamples the feature to the desired resolution and performs element-wise addition with features from previous stages. Specifically, the FPN takes the output feature maps from the last two stages of the backbone as the input and projects them to the same number of channels as SAM, followed by upsampling and addition as standard FPN operations.

We explore various efficient backbones following this paradigm, including ViT-based~\citep{tiny_vit}, CNN-based~\citep{rep_vit}, and hybrid networks~\citep{efficient_vit}. However, we find there is always a considerable performance gap. Training with a longer schedule or using distillation losses~\citep{liu2019structured,shu2021channel} designed for dense prediction tasks does not show an obvious improvement. Therefore, we further propose to consider prompts during distillation to provide task-specific guidance.

\begin{figure*}
    \centering
    \includegraphics[width=1.0\textwidth]{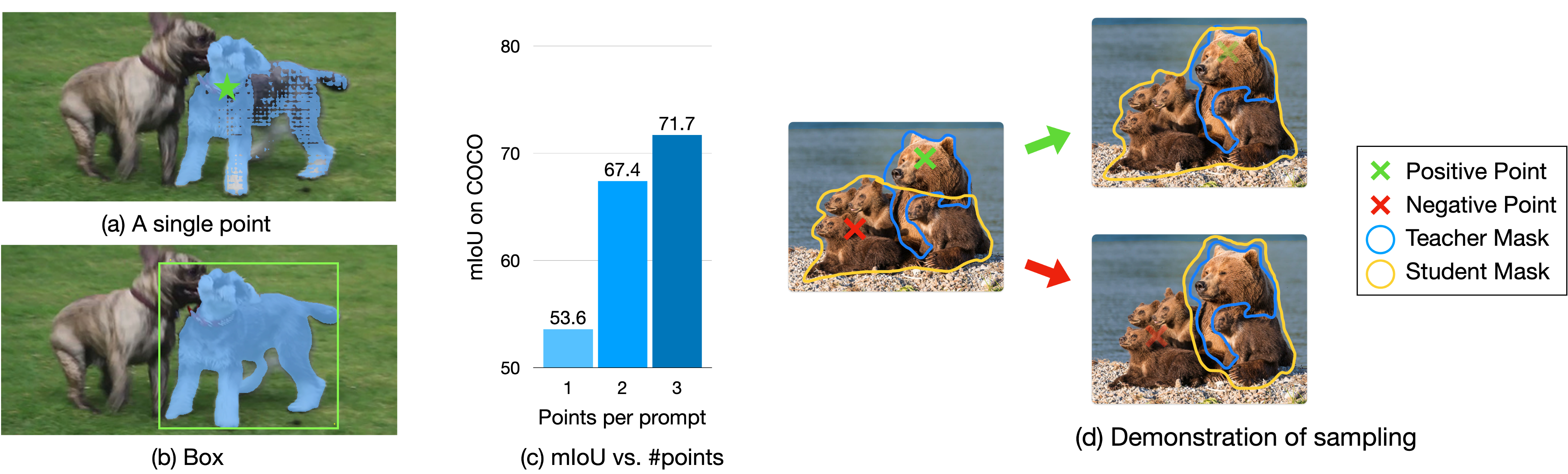}
    \caption{Figures (a-c) show the behaviors of the SAM with different prompts. When prompted with ambiguous prompts, \eg, a point, SAM yields suboptimal results. More informative prompts solve the problem. In (d), we demonstrate the sampling process of prompt-in-the-loop KD. In particular, during distillation, we randomly sample new positive/negative point prompts from the false negative/false positive areas in a loop, so that the student dynamically focuses on those regions.}
    \label{fig:sam_prompt}
\vspace{-5pt}
\end{figure*}

\subsubsection{Prompt-In-the-Loop Knowledge Distillation}
\label{sec:prompt-kd}
Since the original SAM decoder is lightweight enough, taking up only 0.6\% of the total parameters of SAM, we retain its architecture so that the pre-trained weights can be inherited. Here, we re-visit the mask decoder of SAM. It is a two-stream bi-directional transformer, with output feature map $f$ from the image encoder as the input of one stream and sparse prompt embeddings $p$ concatenating with four mask tokens $m$ and an IoU token $\mathbf{c}$ as the input of the other stream\footnote{We ignore the dense prompts and focus only on point and box prompts.}. The sparse prompt can be any combination of positive/negative points and a single box, indicating the user's object of interest.

With these inputs, there are many potential distillation targets for teacher and student models to align with, including the refined feature map, mask/IoU tokens, cross attention between two streams of the inputs, and output mask logits. Through empirical studies, we find that supervising the student with teacher mask output as the ground truth is the most effective. More details are provided in the appendix. We formulate the decoder loss as follows:
\begin{align}
    L_d = L_\text{mask}(\phi(T_{dec}(f_t, p, m, \mathbf{c})), S_{dec}(f_s, p, m, \mathbf{c})),
\end{align}
where $\phi(\cdot)$ is binary thresholding, $f_t$ and $f_s$ denote the feature from the teacher and student image encoder, respectively. The teacher and student share the same set of $p$, $m$, and $\mathbf{c}$, which are kept frozen during training. We adopt a combination of the Dice loss~\citep{dice} and BCE loss as the mask loss $L_\text{mask}$. Note that we allow the gradient backpropagate to the image encoder so that it is jointly learned.

While the distillation loss is frustratingly simple, the prompt selection for each training iteration needs to be carefully designed. We observe that fine-tuning the mask decoder poses a threat to the zero-shot generalization ability. That is, training with a particular combination of prompts sabotages its capability when inferencing with prompt combinations that are not used during training. For instance, training only with points leads to a significant performance drop when testing with box prompts. While freezing the mask decoder or using LoRA~\citep{lora} on the decoder for regulation mitigates the problem, it also limits the performance upper-bound for prompts observed during training. Meanwhile, we find that even the original SAM outputs unsatisfactory mask predictions when prompted with ambiguous prompts, such as a single point. As shown in Fig.~\ref{fig:sam_prompt}, on the COCO dataset, with the object center as the prompt, the mask mIoU of SAM is only 53.6. Thus, aligning the student's outputs in such circumstances might not be optimal.

\SetKwComment{Comment}{/* }{ */}
\begin{algorithm}[t]
    \caption{Prompt-In-the-Loop Distillation}\label{alg}
    $T_{enc}, S_{enc} \gets \text{SAM / EdgeSAM encoder}$\;
    $T_{dec} \gets \text{SAM decoder}$\;
    $S_{dec} \gets \text{EdgeSAM decoder initialized w/}\ T_{dec}$\;
    $m, \mathbf{c} \gets \text{shared mask / IoU tokens}$\;
    $\mathcal{I},\mathcal{P} \gets \text{images and prompts for training}$\;
    $N, M \gets \text{training steps, prompt sampling loops}$\;
    \For{$i = 1,2,\dots,N$}{
        $f_t,f_s \gets T_{enc}(\mathcal{I}_i),S_{enc}(\mathcal{I}_i)$\;
        $p \gets \text{select the box or point prompt in}\ \mathcal{P}_i$\;
        $m_t,m_s \gets T_{dec}(f_t,p,m,\mathbf{c}),S_{dec}(f_s,p,m,\mathbf{c})$\;
        $L \gets L_{\text{mask}}(m_t, m_s)$\;
        \For{$j = 1,2,\dots,M$}{
            $\hat{p} \gets \text{sample\_in\_disagree}(m_t,m_s)$\;
            $p \gets \hat{p} \text{ appends to}\ p$\;
            $m_t,m_s \gets T_{dec}(f_t,p,m,\mathbf{c}),S_{dec}(f_s,p,m,\mathbf{c})$\;
            $L \gets L + L_{\text{mask}}(m_t, m_s)$\;
        }      
        $S_{enc},S_{dec} \gets \text{SGD model update}$\;
    }
\end{algorithm}

To enhance the efficacy of our distillation process, we introduce a dynamic prompt sampling strategy. This approach is designed to achieve three key objectives: (1) dynamically generate a diverse set of prompt combinations from the initial prompt (be it a box or a point), (2) accurately identify areas within the mask where the student model exhibits inaccuracies, thereby directing its focus to these specific segments, and (3) compel the teacher model, namely the SAM, to produce high-quality masks for more precise guidance.

Drawing inspiration from recent advancements in interactive segmentation methods~\citep{interactive_1,interactive_2}, our strategy involves iteratively sampling new prompts in the loop during the distillation phase. 
As detailed in Algo.~\ref{alg}, we start with an initial prompt. With equal probability, either a box or point prompt, as provided by the SA-1B dataset, is input into the decoders of both the teacher and student models. Subsequently, we identify areas where the mask predictions from the teacher and student diverge. As shown in Fig.~\ref{fig:sam_prompt}, using the teacher’s output as the reference, we uniformly sample new prompts: a positive point in regions marked as false negatives, or a negative point in areas identified as false positives. These newly sampled points are then amalgamated with the existing prompts for the subsequent iteration of decoding. Note that we cannot sample new box prompts from the disagreement area because the original SAM only supports up to one box for each object of interest. It is important to note that each prompt leads to four mask predictions at varying levels of granularity. In our analysis, the disagreement is calculated specifically between the teacher mask with the highest IoU score and its corresponding student mask.

In summary, our findings reveal that rather than relying on distillation with losses dedicated to dense prediction or query-based approaches, dynamically feeding appropriate prompt combinations into the mask decoder during the distillation process proves to be more effective. Our prompt-in-the-loop distillation method prioritizes the strategic use of prompts to enhance the learning process. In the experimental section, we present detailed discussions through comprehensive ablation studies.

\subsubsection{Granularity Priors}
\label{sec:rpn}
Since SA-1B is a class-agnostic, multi-grained, automatically labeled dataset, its annotation distribution can be very different from that of the datasets that are intensively labeled by human labor, such as COCO. Therefore, with ambiguous prompts, such as a single point, it is hard for SAM to determine the desired output granularity. Meanwhile, as shown in Fig.~\ref{fig:sam_prompt}, with box prompts, SAM can easily pinpoint the target granularity. In addition, compared to iteratively clicking or interacting with the box, there are many circumstances and applications on smartphones that a single click is favored, such as click-and-drag. Therefore, we propose a simple and efficient module that explicitly embeds the granularity priors of certain datasets and can be optionally turned off if the original behavior of SAM is preferred.

With the image encoder staying frozen, we build a lightweight region proposal network (RPN)~\citep{faster_rcnn} on top of it, which consists of a feature pyramid network (FPN)~\citep{fpn} and a shared detection head. For efficiency, we follow the design proposed by EfficientDet~\citep{efficientdet}. The RPN is trained on a specific dataset, \eg, COCO~\citep{coco}, to capture its granularity prior. During inference, we merge the proposal boxes whose centers are K nearest neighbors of the point prompts weighted by their confidence scores. Finally, we combine the merged box with the point input together as the prompt that inputs to the mask decoder.

\subsection{Training and Application}
\label{sec:train_app}
\noindent\textbf{Training Pipeline.}
We divide the training into three stages. In the first stage, we perform encoder-only knowledge distillation on 1\% SA-1B images following MobileSAM~\citep{mobile_sam}. In the second stage, we apply prompt-in-the-loop distillation on the same image set but with point and box prompts as part of the inputs. In the final stage, which is optional, we freeze modules other than the lightweight RPN and train with the class-agnostic ground-truth boxes with the commonly used focal loss~\citep{focal_loss} and Huber loss~\citep{huber_loss}.

\noindent\textbf{Inference and On-Device Demo.}
EdgeSAM is capable of initiating inference with either point or box prompts, and, akin to SAM, it can progressively incorporate additional points for further refinement. To show the practical utility of EdgeSAM, in \textbf{demo.mov}, we provide a live demonstration of running EdgeSAM on an iPhone. We develop an iOS application and convert the trained EdgeSAM model to a Core ML model serving as the backend (we use the coremltools~\citep{coremltools} for this purpose). The demo video also showcases the qualitative results of applying EdgeSAM to challenging scenes including crowded tiny objects, night-time conditions, different domains, \etc. Our demonstration device is an iPhone 13 running on iOS 17.1. 

\section{Experiments}
\label{sec:exp}

\noindent\textbf{Efficiency.} We measure the model efficiency mainly by throughput per second on a single NVIDIA 2080 Ti/3090 and an iPhone 14 as it reflects the real-world latency. We also report FLOPs and total parameters for side reference. To compare data efficiency, we show the required training sets of each method.

\noindent\textbf{Accuracy.} We divide the accuracy measurement into three scenarios according to the prompt type and source. (1) We use the ground-truth box as the initial prompt and iteratively add more point prompts in the wrongly segmented area for refinement. (2) Similar to (1) but using the center point of the ground-truth mask as the initial prompt. (3) An external object detector is leveraged to provide box prompts. For (1) and (2), we calculate the mIoU across all instances, and for (3), we report mAP and boundary IoU~\citep{boundary_iou}. For all evaluations, we take the output of the first mask token as the prediction\footnote{Since EfficientSAM only supports outputting multiple masks, we select the mask with the highest score as the final prediction for it.}.

\noindent\textbf{Datasets.} We consider SA-1B~\citep{sam}, COCO~\citep{coco}, and LVIS~\citep{lvis} to evaluate both in-domain performance and zero-shot transferability. For distillation, 1\% images of the whole SA-1B dataset are used. We construct a test set, named SA-1K, from SA-1B. In particular, we randomly select 1,000 images from SA-1B, which are not used during training. For each image, we randomly sample 64 instances to avoid memory overflow. For COCO and LVIS, we use their validation sets.

\begin{table*}[t!]
\centering
\caption{\textbf{Efficiency comparison.} FLOPs are calculated based on the $1024 \times 1024$ input resolution. Numbers denoted by * are referred from MobileSAM~\citep{mobile_sam}. Note that, MobileSAM and EfficientSAM-Ti share the same network architecture.}
\begin{tabular}{l|l|ccc|cc}
    \toprule[0.2em]
    \multirow{2}{*}{Model}  & \multirow{2}{*}{Train Set}    & \multicolumn{3}{c|}{FPS}                      & \multirow{2}{*}{MParam.}      & \multirow{2}{*}{GFLOPs} \\
                            &                               & iPhone 14 & 2080 Ti       & 3090              &                               & \\
    \midrule
    SAM                     & SA-1B                         & OOM       & 4.3           & -                 & 641.1                         & 2734.8 \\
    FastSAM                 & 2\% SA-1B                     & -         & -             & 25.0*             & 68.2                          & 887.6   \\
    MobileSAM               & 1\% SA-1B                     & 4.9       & 103.5         & \f{100.0*}        & 9.8                           & 38.2 \\  
    EfficientSAM-Ti         & SA-1B+IN                      & 4.9       & 103.5         & \f{100.0*}        & 9.8                           & 38.2 \\
    \textbf{EdgeSAM}        & 1\% SA-1B                     & \f{38.7}  & \f{164.3}     & -                 & \f{9.6}                       & \f{22.1} \\
    \textbf{EdgeSAM-RPN}    & 1\% SA-1B                     & 34.1      & 123.9         & -                 & 9.8                           & 22.2 \\
    \bottomrule[0.2em]
\end{tabular}
\label{tab:speed}
\end{table*}
\begin{table*}[t!]
\centering
\caption{\textbf{Performance with GT boxes as prompts.} We report the mIoU across all instances in the test set. \emph{+1 pt.} denotes appending an additional refinement point as the prompt. \textbf{Bold} marks the best while \underline{underline} marks the second best. Since EfficientSAM is trained on the entire SA-1B dataset, we do not evaluate it on SA-1K.}

\begin{tabular}{l|ccc|ccc|ccc}
    \toprule[0.2em]
    \multirow{2}{*}{Method} & \multicolumn{3}{c|}{SA-1K}            & \multicolumn{3}{c|}{COCO}         & \multicolumn{3}{c}{LVIS} \\
                            & Box       & +1 pt.    & +2 pt.        & Box       & +1 pt.    & +2 pt.    & Box       & +1 pt.    & +2 pt.\\
    \midrule
    SAM                     & \f{86.7}  & \f{86.7}  & \f{87.1}      & \f{77.3}  & \s{77.7}  & \s{78.1}  & \f{77.8}  & \f{78.3}  & \f{78.5} \\
    MobileSAM               & 82.0      & 82.4      & 82.7          & 74.4      & 74.8      & 75.1      & 73.1      & 73.7      & 74.0 \\  
    EfficientSAM-Ti         & -         & -         & -             & 75.2      & 76.0      & 76.6      & 74.6      & 75.2      & 75.1 \\
    \textbf{EdgeSAM}        & \s{83.0}  & \s{83.7}  & \s{84.1}      & \s{76.7}  & \f{78.1}  & \f{79.0}  & \s{76.2}  & \s{77.3}  & \s{78.0} \\
    \bottomrule[0.2em]
\end{tabular}

\label{tab:gt_box}
\end{table*}

\subsection{Implementation Details}
\label{sec:impl}
As mentioned in Sec.~\ref{sec:train_app}, we split the training into three stages. In the first stage, we train on 1\% of SA-1B images with encoder-only distillation loss $L_p$. We train the model for 10 epochs with the batch size set to 64. We adopt the AdamW optimizer~\citep{adamw} and set the initial learning rate to 1.25e-2. We adopt the cosine decay schedule, which gradually decreases the learning rate to 6.25e-7. In the second stage, we load the encoder weights trained in stage one and inherit decoder weights from SAM. We then train with the prompt-in-the-loop distillation loss $L_d$ on 1\% SA-1B for 5 epochs. The batch size is 16, and the learning rate decays from 1e-4 to 1e-5. We set the maximum allowed instances per image to 16, and the number of loops for prompt sampling is 1. In the final optional stage for RPN training, we follow the 1x schedule of MMDetection~\citep{mmdetection} on COCO~\citep{coco}. Note that, for both the first and second stages, following TinyViT~\citep{tiny_vit}, we cache output embeddings from the teacher's encoder to disk to minimize the training overhead. In addition, since EdgeSAM is data efficient and only requires 1\% of SA-1B samples, the total training cost is lower than MobileSAM.

\subsection{Quantitative Results}
\label{sec:quant}

\noindent\textbf{Efficiency.} 
In Tab.~\ref{tab:speed}, we compare the FPS of SAM, FastSAM, MobileSAM, EfficientSAM-Ti, and EdgeSAM on both the desktop and mobile platforms. When evaluating with NVIDIA 2080 Ti, we compile the models with ONNX to avoid the overhead of the Python interpreter, and we make sure the GPU is running at 100\% payload. On iPhone 14, we use the Core ML Tools~\citep{coremltools} as the compiler and measure the throughput with the benchmark tools provided by Xcode. As shown in Tab.~\ref{tab:speed}, the proposed EdgeSAM is significantly faster than other methods, especially on the mobile platform. In particular, EdgeSAM is \emph{over 37 times faster} than SAM on NVIDIA 2080 Ti and \emph{over 7 times} faster than MobileSAM and EfficientSAM on iPhone 14. From the benchmark report, we observe that MobileSAM and EfficientSAM are not well optimized by on-device AI accelerators, such as ANE, which explains the vast differences between desktop and mobile platforms.
In addition, EfficientSAM requires a huge training cost including the 400 epochs on ImageNet-1K (IN) and 5 epochs on the entire SA-1B dataset.

\begin{table*}[t!]
\centering
\caption{\textbf{Performance with center points as prompts.} Similar to Tab.~\ref{tab:gt_box} but using the mask center point as the initial prompt.}
\begin{tabular}{l|ccc|ccc|ccc}
    \toprule[0.2em]
    \multirow{2}{*}{Method} & \multicolumn{3}{c|}{SA-1K}            & \multicolumn{3}{c|}{COCO}             &  \multicolumn{3}{c}{LVIS} \\
                            & Center    & +1 pt.    & +2 pt.        & Center        & +1 pt.    & +2 pt.    & Center    & +1 pt.    & +2 pt.\\
    \midrule
    SAM                     & \f{76.5}  & \f{83.4}  & \f{85.1}      & \f{53.6}      & \f{67.4}  & \f{71.7}  & \f{60.5}  & \f{68.1}  & \f{70.7} \\
    MobileSAM               & 64.6      & 73.4      & 76.2          & \s{50.9}      & \s{63.0}  & 66.8      & 52.1      & 59.9      & 63.0 \\  
    EfficientSAM-Ti         & -         & -         & -             & 49.8          & 60.5      & 65.7      & \s{56.4}  & 62.5      & 65.4 \\
    \textbf{EdgeSAM}        & \s{67.5}  & \s{76.1}  & \s{79.0}      & 48.0          & 61.8      & \s{68.7}  & 53.7      & \s{63.4}  & \s{67.7} \\
    \midrule
    \textbf{EdgeSAM-RPN}    &           &           &               & 54.3          &           &           &           &          & \\
    \bottomrule[0.2em]
\end{tabular}
\label{tab:center_pt}
\end{table*}
\begin{table*}[t!]
\centering
\caption{\textbf{Performance with boxes from an external object detector as prompts.} We report the mask mAP and boundary IoU on the COCO dataset. The box mAPs of Detic and ViTDet-H are 47.4 and 58.7 respectively.}
\begin{tabular}{l|ccccc|cccc|c|c}
    \toprule[0.2em]
    \multirow{2}{*}{Method} & \multicolumn{5}{c|}{Detic}                                & \multicolumn{4}{c|}{ViTDet-H}                 & \multirow{2}{*}{Train Set}    & \multirow{2}{*}{FPS} \\
                            & AP        & \aps      & \apm      & \apl      & BIoU      & AP        & \aps      & \apm      & \apl      &                               & \\
    \midrule
    SAM                     & \f{38.8}  & \f{26.9}  & \f{44.1}  & \f{50.3}  & \f{26.8}  & \f{46.1}  & \f{33.6}  & \f{51.9}  & \f{57.7}  & SA-1B                         & 4.3   \\
    FastSAM                 & -         & -         & -         & -         & -         & 37.9      & 23.9      & 43.4      & 50.0      & 2\% SA-1B                     & $<$103.5 \\
    MobileSAM               & 33.1      & 21.7      & 37.8      & 44.8      & 20.2      & 39.4      & 26.9      & 44.4      & 52.2      & 1\% SA-1B                     & \s{103.5} \\  
    EfficientSAM-Ti         & -         & -         & -         & -         & -         & \s{42.3}  & 26.7      & 46.2      & \s{57.4}  & SA-1B+IN                      & \s{103.5} \\
    \textbf{EdgeSAM}        & \s{35.2}  & \s{23.5}  & \s{40.3}  & \s{46.6}  & \s{22.5}  & 42.2      & \s{29.6}  & \s{47.6}  & 53.9      & 1\% SA-1B                     & \f{164.3} \\
    \bottomrule[0.2em]
\end{tabular}
\label{tab:det_box}
\end{table*}

\noindent\textbf{Prompting with GT Boxes.}
Despite being extremely lightweight, EdgeSAM can generate accurate segmentation masks. As shown in Tab.~\ref{tab:gt_box}, the proposed EdgeSAM consistently outperforms MobileSAM and EfficientSAM across a wide range of prompt combinations and datasets, demonstrating its effectiveness for real-world applications, including zero-shot transferability and iterative refinement. To our surprise, on the COCO dataset, with one or two points serving as the refinement prompts, EdgeSAM even surpasses the SAM.

\noindent\textbf{Prompting with Center Points.}
Similar to prompting with GT boxes, EdgeSAM shows a clear advantage over MobileSAM and EfficientSAM in most cases, as shown in Tab.~\ref{tab:center_pt}. However, as we have discussed in Sec.~\ref{sec:rpn}, with ambiguous prompts, such as a single point, the original SAM does not always produce masks in the desired granularity. In addition, EdgeSAM is trained on the SA-1B dataset, whose granularity distribution is very different from that of COCO. Therefore, with the proposed task-aware distillation, EdgeSAM is more likely to fit the SAM granularity distribution when handling ambiguous prompts than MobileSAM. To explicitly capture such a granularity prior, we propose a lightweight RPN, which works as expected and makes the single point performance on COCO rise from 48.0 to 54.3. As we do not observe a granularity degradation on other datasets except for COCO, we refrain from applying RPN on them to keep the model simple and efficient. Besides, when compared with EfficientSAM, we would like to highlight that it is trained on a 100 times larger dataset than EdgeSAM and uses a larger backbone model. Later in the ablation studies (Sec.~\ref{sec:abla}), we show that EdgeSAM can also benefit from larger train sets and models.

\begin{table*}[t!]
\centering
\caption{\textbf{Ablation studies.} If not explicitly stated, we report the mIoUs on SA-1K and FPS on a single NVIDIA 2080 Ti. COCO mAP and boundary IoU~\citep{boundary_iou} are obtained with the Detic~\citep{detic} object detector providing box prompts and class labels (box mAP 47.4). \textbf{Bold} font indicates the best number and the default configuration we adopt.}
\vspace{-3pt}

\begin{subtable}[c]{0.6\textwidth} 
\centering
\caption{\textbf{Effectiveness of prompt-in-the-loop knowledge distillation (KD).}}
\vspace{-2pt}
\label{tab:abl-prompt-kd}
\begin{smalltable}{l|ccc|ccc}
    \toprule[0.2em]
    Method                      & Box       & +1 Pt.    & +2 Pt.    & Center    & +1 Pt.    & +2 Pt.    \\
    \midrule
    encoder-only KD             & 82.0      & 82.4      & 82.8      & 64.6      & 73.3      & 76.2      \\
    + prompt-in-the-loop KD     & \f{83.0}  & \f{83.7}  & \f{84.1}  & \f{67.5}  & \f{76.1}  & \f{79.0}  \\
    \bottomrule[0.2em]
\end{smalltable}
\end{subtable}
\quad
\begin{subtable}[c]{0.35\textwidth} 
\centering
\caption{\textbf{Speed-performance trade-off of the RPN for granularity priors.} Evaluated on COCO with center point as the prompt.}\label{tab:abl-rpn}
\vspace{-2pt}
\begin{smalltable}{l|cccc}
    \toprule[0.2em]
    Method                  & mIoU      & 2080 Ti       & iPhone 14   \\
    \midrule
    w/o RPN                 & 48.0      & \f{164.3}     & \f{38.7}    \\
    w/ RPN                  & \f{54.3}  & 123.9         & 34.1        \\
    \bottomrule[0.2em]
\end{smalltable}
\end{subtable}

\begin{subtable}[c]{0.5\textwidth} 
\centering
\caption{\textbf{Choice of the backbone.} We apply encoder-only KD for this ablation.}\label{tab:abl-backbone}
\vspace{-2pt}
\begin{smalltable}{l|ccccc}
    \toprule[0.2em]
    Method                  & Res. Align        & Type      & Box       & Center    & FPS    \\
    \midrule
    TinyViT-5M              & \multirow{3}{*}{\makecell{Remove\\Downsample}}   
                                                & ViT       & 82.0      & 64.6      & 103.5   \\
    EfficientViT-B1         &                   & Hybrid    & 81.6      & 63.7      & 117.0   \\
    RepViT-M1               &                   & CNN       & \f{82.1}  & \f{64.9}  & 155.7   \\
    \midrule
    TinyViT-5M              & \multirow{3}{*}{FPN}
                                                & ViT       & 81.6      & 63.7      & 114.2     \\
    EfficientViT-B1         &                   & Hybrid    & 80.7      & 60.9      & 159.9     \\
    RepViT-M1               &                   & CNN       & 82.0      & 64.6      & \f{164.3} \\
    \bottomrule[0.2em]
\end{smalltable}
\end{subtable}
\quad
\begin{subtable}[c]{0.4\textwidth} 
\centering
\caption{\textbf{Generalizability of prompt-in-the-loop KD.}}\label{tab:abl-generalizability}
\vspace{-2pt}
\begin{smalltable}{l|cccc}
    \toprule[0.2em]
        Backbone                & Our KD                & Box       & Center    & FPS    \\
        \midrule
        TinyViT-5M              & \xmark                & 81.6      & 63.7      & 114.2  \\
        EfficientViT-B1         & \xmark                & 80.7      & 60.9      & 159.9  \\
        RepViT-M1               & \xmark                & 82.0      & 64.6      & 164.3  \\
        \midrule
        TinyViT-5M              & \cmark                & 83.5      & 68.0      & 114.2  \\
        EfficientViT-B1         & \cmark                & 82.7      & 65.8      & 159.9  \\
        RepViT-M1               & \cmark                & 83.0      & 67.5      & 164.3  \\
        \bottomrule
\end{smalltable}
\end{subtable}

\begin{subtable}[c]{0.55\textwidth} 
\centering
\caption{\textbf{Train with more data.}}\label{tab:abl-more-data}
\vspace{-2pt}
\begin{smalltable}{l|l|cccc|c}
    \toprule[0.2em]
        \multirow{2}{*}{Method}     & \multirow{2}{*}{Train Set}    &\multicolumn{4}{c|}{ViTDet-H}              & \multirow{2}{*}{FPS} \\
                                    &                               & AP        & \aps     & \apm     & \apl    & \\
        \midrule    
        FastSAM                     & 2\% SA-1B                     & 37.9      & 23.9      & 43.4      & 50.0  & -     \\
        MobileSAM                   & 1\% SA-1B                     & 39.4      & 26.9      & 44.4      & 52.2  & 103.5 \\
        EfficientSAM-Ti             & SA-1B+IN                      & 42.3      & 26.7      & 46.2      & 57.4  & 103.5 \\
        \midrule
        EdgeSAM                     & 1\% SA-1B                     & 42.2      & 29.6     & 47.6     & 53.9    & 164.3 \\
        EdgeSAM-3x                  & 3\% SA-1B                     & 42.7      & 30.0     & 48.6     & 54.5    & 164.3 \\
        EdgeSAM-10x                 & 10\% SA-1B                    & 43.0      & 30.3     & 48.9     & 55.1    & 164.3 \\
    \bottomrule[0.2em]
\end{smalltable}
\end{subtable}
\begin{subtable}[c]{0.38\textwidth} 
\centering
\caption{\textbf{Frozen modules and LoRA.} F denotes freezing and L means applying LoRA during training.}\label{tab:abl-freeze}
\vspace{-2pt}
\begin{smalltable}{cc|cc|cc}
    \toprule[0.2em]
    \multirow{2}{*}{Enc.}       & \multirow{2}{*}{Dec.}     & \multicolumn{2}{c|}{SA-1K}    & \multicolumn{2}{c}{COCO}          \\
                                &                           & Box       & Center            & Box       & Center                \\
    \midrule
    F                           & -                         & 82.3      & 65.7              & 76.4      & 47.4                  \\
    -                           & F                         & \f{83.5}  & \f{67.7}          & 75.9      & \f{48.3}              \\
    -                           & L                         & 82.9      & 66.8              & 74.9      & 46.8                  \\
    -                           & -                         & 83.1      & 67.4              & \f{76.7}  & 48.2                  \\
    \bottomrule[0.2em]
\end{smalltable}
\end{subtable}

\begin{subtable}[c]{0.45\textwidth} 
\centering
\caption{\textbf{Number of prompt sampling loops.}}\label{tab:abl-loop}
\vspace{-2pt}
\begin{smalltable}{c|ccc|ccc}
    \toprule[0.2em]
    Loops               & Box       & +1 Pt.    & +2 Pt.    & Center    & +1 Pt.    & +2 Pt. \\
    \midrule
    0                   & \f{83.1}  & 83.2      & 83.4      & 67.4      & 71.1      & 72.4   \\
    \f{1}               & 83.0      & \f{83.7}  & \f{84.1}  & \f{67.5}  & \f{76.1}  & 79.0  \\
    2                   & 82.9      & \f{83.7}  & \f{84.1}  & 67.3      & \f{76.1}  & \f{79.2} \\
    \bottomrule[0.2em]
\end{smalltable}
\end{subtable}
\begin{subtable}[c]{0.45\textwidth} 
\centering
\caption{\textbf{Number of prompts.} Mask decoder is frozen.}\label{tab:supp-abl-num-prompt}
\vspace{-2pt}
\begin{smalltable}{l|cc|cc}
    \toprule[0.2em]
    \multirow{2}{*}{\#Prompt}       & \multicolumn{2}{c|}{SA-1K}    & \multicolumn{2}{c}{COCO} \\
                                    & Box       & Center            & AP        & BIoU \\
    \midrule
    8                               & \f{83.6}  & 67.6              & 34.3      & 21.7 \\
    \textbf{16}                     & 83.5      & \f{67.7}          & \f{34.6}  & \f{21.9} \\
    32                              & \f{83.6}  & 67.6              & 34.3      & 21.6 \\
    \bottomrule[0.2em]
\end{smalltable}
\end{subtable}

\begin{subtable}[c]{0.3\textwidth} 
\centering
\caption{\textbf{Encoder-only KD.}}\label{tab:abl-enc-kd}
\vspace{-2pt}
\begin{smalltable}{l|cc}
    \toprule[0.2em]
    Method                                  & Box       & Center    \\
    \midrule
    Baseline                                & \f{82.0}  & 64.6      \\
    20 Epochs                               & 81.9      & \f{65.9}  \\
    w/ Structured                           & 81.6      & 65.2      \\
    w/ Channel-wise                         & 81.8      & 64.9      \\
    \bottomrule[0.2em]
\end{smalltable}
\end{subtable}
\begin{subtable}[c]{0.4\textwidth} 
\centering
\caption{\textbf{Decoder losses.}}\label{tab:supp-abl-dec-loss}
\vspace{-2pt}
\begin{smalltable}{l|cc|cc}
    \toprule[0.2em]
    \multirow{2}{*}{Method}     & \multicolumn{2}{c|}{SA-1K}    & \multicolumn{2}{c}{COCO} \\
                                & Box       & Center            & AP        & BIoU \\
    \midrule
    \textbf{Mask}               & \f{83.1}  & \f{67.4}          & \f{35.1}  & \f{22.4} \\
    Mask + Attn.                & 82.8      & 67.0              & 35.0      & 22.3 \\
    Mask + IoU                  & 82.5      & 67.1              & 34.9      & 22.2 \\
    Img. Feat.                  & 71.6      & 55.0              & 30.2      & 17.2 \\
    \bottomrule[0.2em]
\end{smalltable}
\end{subtable}
\begin{subtable}[c]{0.2\textwidth} 
\centering
\caption{\textbf{Train stages.}}\label{tab:abl-stage}
\vspace{-2pt}
\begin{smalltable}{l|cc}
    \toprule[0.2em]
                        & Box       & Center   \\
    \midrule
    Joint               & 81.2      & 63.2      \\
    \f{2-stage}         & \f{83.1}  & \f{67.4}  \\
    \bottomrule[0.2em]
\end{smalltable}
\end{subtable}

\begin{subtable}[c]{1.0\textwidth} 
\centering
\caption{\textbf{Training datasets.} SA-1B* means 1\% SA-1B.} \label{tab:supp-abl-dataset}
\vspace{-2pt}
\begin{smalltable}{l|cc|cc|cc}
    \toprule[0.2em]
    \multirow{2}{*}{Dataset}    & \multicolumn{2}{c|}{SA-1K}    & \multicolumn{2}{c|}{COCO}     & \multicolumn{2}{c}{LVIS}  \\
                                & Box       & Center            & Box      & Center             & Box       & Center   \\
    \midrule
    \textbf{SA-1B*}             & \f{83.1}  & \f{67.4}          & \f{76.7} & 48.2               & \f{76.1}  & 53.5      \\
    COCO                        & 81.3      & 63.1              & 76.1     & \f{50.8}           & 74.7      & 54.2      \\
    LVIS                        & 82.4      & 63.6              & 75.7     & 50.2               & 75.4      & \f{54.9}  \\
    \bottomrule[0.2em]
\end{smalltable}
\end{subtable}

\end{table*}

\noindent\textbf{Prompting with External Object Detectors.}
We finally combine all the SAM variants with an external object detector (We use Detic~\citep{detic} from MMDetection~\citep{mmdetection} and ViTDet-H~\citep{vitdet} from Detectron2~\citep{detectron2}) and evaluate their mAP performance on COCO. This setting examines a practical use case where automatic segmentation is preferred over interactive segmentation, \eg, generating the mask within the tracking box. We use an object detection framework to generate bounding boxes and associated class labels, ensuring that the observed variations in mAP are attributed solely to the quality of the masks generated. This approach also evaluates the resilience of our model in handling imprecise bounding box inputs.
We also report boundary IoU~\citep{boundary_iou} to reflect the boundary accuracy. Tab.~\ref{tab:det_box} shows that EdgeSAM outperforms MobileSAM by considerable margins and is on par with EfficientSAM with smaller training and inference costs.
However, it still lags clearly behind SAM. This performance gap may reflect not only inherent limitations in model capacity but also a consequence of training exclusively with ground-truth boxes, which could lead to discrepancies during inference. Addressing these differences and exploring their implications is a promising direction for future research.

\subsection{Ablation Studies}
\label{sec:abla}

\noindent\textbf{Effectiveness of Each Proposed Component.} 
We investigate the effectiveness of the introduced techniques, namely, prompt-in-the-loop knowledge distillation (prompt-KD) and the lightweight Region Proposal Network (RPN) that incorporates granularity priors. 
As depicted in Tab.~\ref{tab:abl-prompt-kd}, prompt-KD demonstrates a notable enhancement in performance over encoder-only knowledge distillation, particularly when complemented with additional refinement points. Prompt-KD's advantage lies in its provision of task-specific supervision, which is more explicit and targeted than the general guidance offered by encoder-only KD. 
In addition, its strategy of dynamically generating new prompts in inaccurately segmented areas places more focus on these regions, creating diverse prompt combinations in the process. 
Furthermore, as shown in Tab.~\ref{tab:abl-rpn}, the lightweight RPN operates effectively at a manageable computational cost. It is important to note that during the RPN's training phase, we freeze the other components, including the backbone network. This approach allows for the RPN to be dynamically deactivated during inference, ensuring that the model's generalization capabilities remain intact.

\noindent\textbf{Choice of the Backbone.}
As mentioned in Sec.~\ref{sec:encoder-kd}, the output resolution of SAM and that of efficient backbones are misaligned. MobileSAM proposed to remove the downsampling operations in the last several stages, while we experiment with fusing low-resolution feature maps with higher ones through an FPN~\citep{fpn}. In addition, we compare the performance-speed trade-offs across efficient backbones following the ViT-based~\citep{tiny_vit}, CNN-based~\citep{rep_vit}, and hybrid~\citep{efficient_vit} designs. As shown in Tab.~\ref{tab:abl-backbone}, the purely CNN-based RepViT-M1 with FPN for resolution alignment achieves the best balance. Moreover, as many on-device AI accelerators, such as ANE, are highly optimized for CNNs, the speed gap becomes larger when deployed on edge devices. For example, on iPhone 14, with FPN for resolution alignment, RepViT-M1 only takes 14 ms to encode a $1024\times 1024$ input, which is 14x and 4x faster than TinyViT-5M and EfficientViT-B1 respectively. Therefore, our EdgeSAM adopts it as the image encoder.

\noindent\textbf{Generalizability of the Proposed Prompt-in-the-Loop KD.}
While we opt for RepViT given it is more optimized for current mobile platforms (empirically proved by the previous ablation), the proposed prompt-in-the-loop knowledge distillation method is not tied to a specific backbone. Therefore, in Tab.~\ref{tab:abl-generalizability}, we show that with the proposed KD method, the performance of the resulting model is consistently and significantly improved across TinyViT, EfficientViT, and RepViT without any additional cost during inference, which demonstrates the generalizability of our method. Note that, RepViT-M1 without our prompt-in-the-loop KD is essentially RepViT-SAM~\citep{repvit-sam}. Therefore, EdgeSAM outperforms RepViT-SAM by 1.0 and 2.9 mIoUs with box and center point as the prompt, respectively \footnote{The original RepViT-SAM adopts RepViT-M2.3 as the encoder while here we compare the performance with the RepViT-M1 backbone.}.

\noindent\textbf{Train with More Data.}
By default, EdgeSAM is trained with 1\% data of SA-1B. However, we observe that FastSAM and EfficientSAM leverage more training samples. Thus, we further train EdgeSAM with more data while keeping other hyper-parameters fixed to inspect whether EdgeSAM can benefit from more training samples. Tab.~\ref{tab:abl-more-data} shows a clear advantage of using more data, \eg, AP is boosted from 42.2 to 43.0 by using 10\% SA-1B, which in the same time surpasses EfficientSAM by 0.7 AP.

\noindent\textbf{Frozen Modules and LoRA.}
In Tab.~\ref{tab:abl-freeze}, we test several configurations regarding module freezing and LoRA~\citep{lora}. In particular, LoRA is applied to the query and value projection layers of the attention blocks in the decoder. Results show that freezing the decoder during distillation yields the best in-domain accuracy while fine-tuning all the modules generalizes better on other datasets. To benefit more practical settings, we opt to fine-tune both the encoder and decoder. 

\noindent\textbf{Number of Prompt Sampling Loops.}
In Tab.~\ref{tab:abl-loop}, we find that one additional loop achieves the best cost-performance trade-off, and we conjecture that with one correction, the student is already doing better enough to match the teacher, that is, the prediction difference between student and teacher does not leave enough sampling area to generate new prompts from. Thus we set the loop number to be 1 by default. Note that when setting the number of loops to 0, the decoder is still inferred once and gets trained, which is different from encoder-only KD where the decoder is not involved during training.

\noindent\textbf{Number of Prompts.}
Given that each sample of SA-1B contains around 100 instances, training all of them in a single batch causes VRAM overflow. Therefore, we randomly sample a subset for each iteration. In Tab.~\ref{tab:supp-abl-num-prompt}, we study the impact of the number of prompts for sampling and find that 16 prompts achieve the best trade-off. Thus, we set it as the default.

\noindent\textbf{Encoder-Only KD.}
In Tab.~\ref{tab:abl-enc-kd}, we explore several ways to improve the performance of encoder-only distillation, including training for a longer schedule and applying knowledge distillation losses designed for dense prediction tasks~\citep{liu2019structured,shu2021channel}. While they boost the performance of point prompts, that of box prompts slightly drops, indicating an upper-bound performance of encoder-only distillation. This motivates us to explore prompt-aware distillation.

\noindent\textbf{Training Losses for Decoder.}
By default, we supervise the student with teacher mask output as the ground truth is the most effective for decoder loss. Here, we provide more details of our empirical findings. In particular, we compare various configurations, including (1) combining mask loss with aligning attention maps from the teacher and student decoder; (2) combining mask loss with aligning the IoU predictions between teacher and student; (3) aligning the output feature maps between the teacher and student decoder. As shown in Tab.~\ref{tab:supp-abl-dec-loss}, the plain mask loss supervision yields the best performance. Thus, instead of exploring different loss combinations, we study the strategy to select the proper prompts during distillation and propose prompt-in-the-loop distillation.

\noindent\textbf{Train Stages.}
Due to the optimal optimization settings being different for encoder-only distillation and prompt-in-the-loop distillation, we find joint-training yields sub-optimal results compared to training in two stages (see Tab.~\ref{tab:abl-stage}). Therefore, we propose to train them sequentially. 

\noindent\textbf{Training Datasets.}
In our default setting, we train EdgeSAM with 1\% images of the SA-1B~\citep{sam} dataset. In Tab.~\ref{tab:supp-abl-dataset}, we study the impact of training datasets on transferability. Specifically, we keep the model and training pipeline the same while switching the training dataset among 1\% SA-1B, COCO~\citep{coco}, and LVIS~\citep{lvis}. Note that all datasets contain a similar number of images (approximately 110K). Results show that training with SA-1B archives the best performance across all test sets with box prompts, while with point prompts, training and testing in the same domain yield the best result. This suggests that with informative prompts, such as boxes, distillation on the SA-1B dataset is sufficient for zero-shot transfer. However, with ambiguous prompts, such as a single point, efforts must be made to capture the granularity priors of the testing datasets. This motivates us to leverage an additional RPN to capture such priors explicitly.

\begin{figure*}
    \centering
    \includegraphics[width=1.0\textwidth]{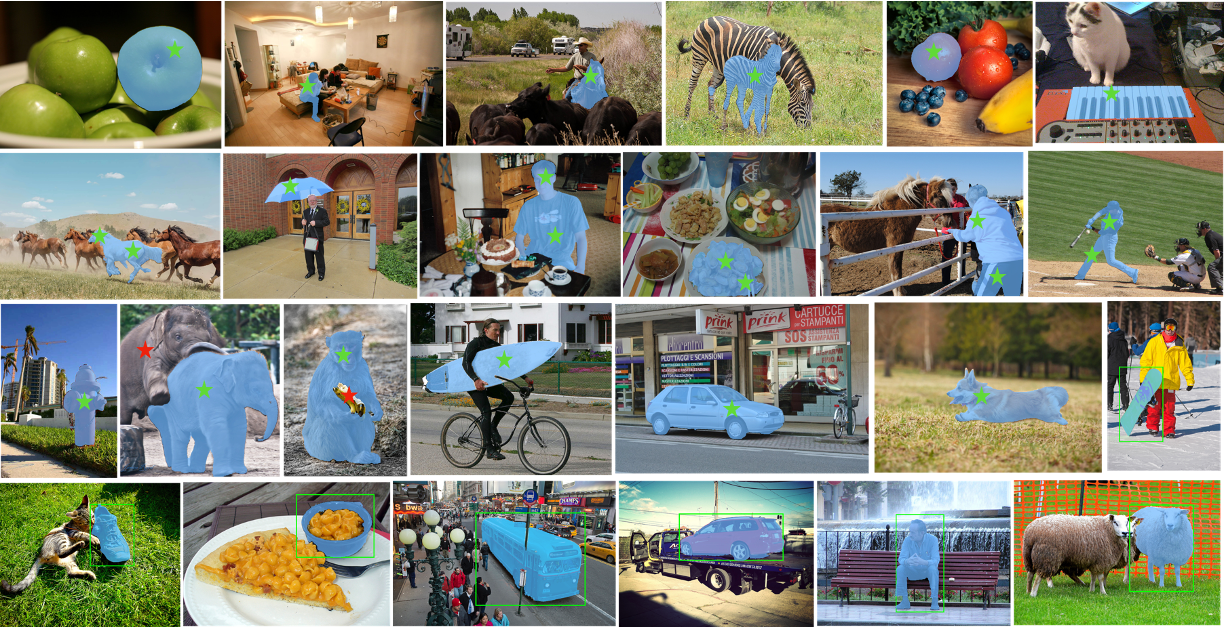}
    \caption{\textbf{Qualitative results of EdgeSAM with point and box prompts.} The green and red stars indicate the positive and negative points respectively.}
    \label{fig:vis}
    \vspace{-10pt}
\end{figure*}

\begin{figure*}
    \centering
    \includegraphics[width=0.95\textwidth]{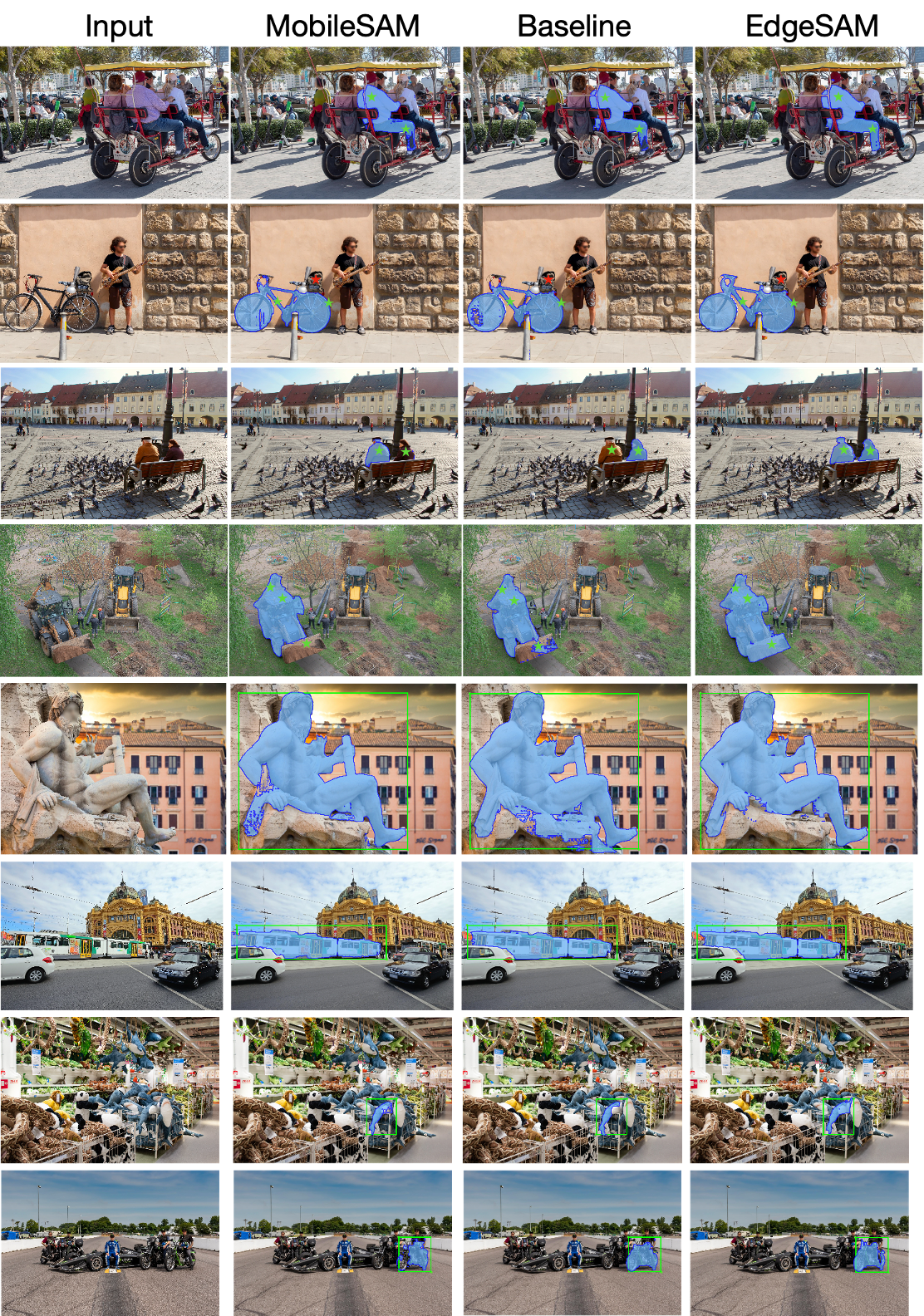}
    \caption{We compare the visual results of applying prompt-in-the-loop knowledge distillation or not to demonstrate its effectiveness. The baseline follows the same network architecture of EdgeSAM but adopts MobileSAM's encoder-only distillation method.}
    \label{fig:supp-compare}
\end{figure*}
\begin{figure*}[t]
    \centering
    \includegraphics[width=0.95\textwidth]{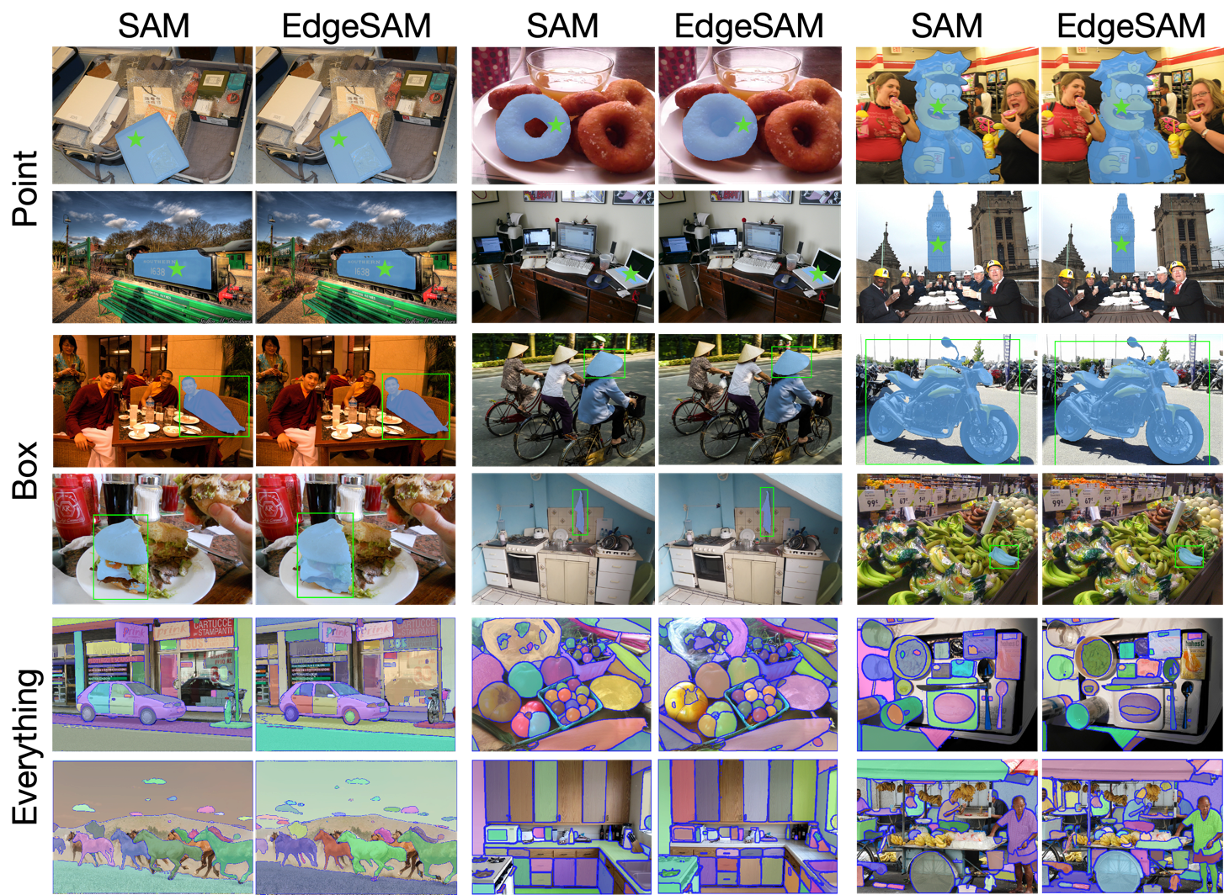}
    \caption{\textbf{Qualitative results of EdgeSAM and SAM under point, box, and everything mode.}}
    \label{fig:supp-vis}
\end{figure*}

\subsection{Qualitative Results}
\label{sec:qual}
In Fig.~\ref{fig:vis}, we visualize the mask quality of EdgeSAM with one point, two points, and one box as the prompts, respectively. 
Note that \textit{none} of the images are from the SA-1B dataset that EdgeSAM is trained on, demonstrating its zero-shot generalization ability. 

\subsection{Effectiveness of the Prompt-in-the-Loop Knowledge Distillation}
\label{sec:supp-compare}
In Fig.~\ref{fig:supp-compare}, we compare EdgeSAM with MobileSAM with both points and boxes as prompts. Since EdgeSAM and MobileSAM adopt different backbones, we add a baseline that follows the same network architecture as EdgeSAM but is trained with MobileSAM's encoder-only distillation method. Compared with task-agnostic KD methods, models trained with prompt-KD follow the user's intent more precisely, \eg, MobileSAM and baseline fail to segment both persons on the bench. Besides, prompt-KD involves implicit hard example mining by iteratively sampling new prompts in the wrongly segmented area, which leads to tighter boundaries and more accurate results in difficult regions, \eg, the sculpture and bicycle in Fig.~\ref{fig:supp-compare}. These results align well with our assumption that task-agnostic encoder distillation fails to capture the full knowledge embodied in SAM and involving prompts in the loop during distillation helps capture the intricate dynamics between user input and mask generation.

\subsection{Qualitative Results Compared with SAM}
\label{sec:supp-vis}
In Fig.~\ref{fig:supp-vis}, we provide more qualitative results of EdgeSAM and compare it with the original SAM under single point, box, and everything mode. Note that since EdgeSAM follows the same encoder-decoder architecture as SAM, everything mode is costly. Specifically, in everything mode, points in a $32 \times 32$ grid are fed to the decoder, which results in 1,024 times decoder inference. As EdgeSAM is built for inference-speed-sensitive devices, we do not advocate using everything mode of EdgeSAM in practice.  Therefore, we provide visualization examples mainly to discuss its behavior compared to SAM. As expected, the overall mask quality of EdgeSAM is inferior to SAM, but they are comparable, especially for box prompts. One typical failure case for EdgeSAM is mis-segmenting the holes within an object, for instance, the donut example in Fig.~\ref{fig:supp-vis} (row 1, col 4). We argue this might be due to the lack of training samples. But in actual use, one can add a negative point in the hole area to explicitly force EdgeSAM to exclude such regions. Moreover, only masks with confident scores over a certain threshold are kept in everything mode. Thus, regions without masks indicate unsatisfactory mask predictions. Same as SAM, EdgeSAM covers most objects in the scene, demonstrating its capability.

\section{Conclusion and Discussion}
\label{sec:conclusion}

In this paper, we present EdgeSAM, the first SAM variant that runs in real-time on edge devices. We achieve this by distilling SAM into a lightweight CNN-based architecture. Our preliminary experiments show that existing distillation schemes, which only involve the image encoder, thus being task-agnostic, fail to reveal the full knowledge spectrum of SAM to the student model. Thus, we propose a prompt-in-the-loop knowledge distillation method that considers both the encoder and decoder of SAM and provides task-specific supervision signals. Experiments across SA-1B, COCO, and LVIS under various settings demonstrate the effectiveness of EdgeSAM in terms of efficiency and accuracy compared with SAM and other accelerated variants. 

Despite being able to run EdgeSAM at a decent speed on edge devices, we still have several research directions, that could potentially provide boosts and is orthogonal to our contributions in this paper, left to be investigated, including quantization, model pruning, on-device optimization, mixed-precision inference, \etc. Besides, we haven't applied any augmentation during training, thus, proper data or prompt augmentation might also be promising directions. We hope EdgeSAM will encourage more real-world applications with its on-device interactive segmentation capability and will continue working on further improvements.

\section{Data Availability Statements}
The datasets used in this study are publicly available as follows:
\begin{itemize}
    \item SA-1B: \href{https://ai.meta.com/datasets/segment-anything-downloads/}{https://ai.meta.com/datasets/segment-anything-downloads/}
    \item COCO: \href{https://cocodataset.org/\#download}{https://cocodataset.org/\#download}
    \item LVIS: \href{https://www.lvisdataset.org/dataset}{https://www.lvisdataset.org/dataset}
\end{itemize}

{\small
\bibliographystyle{spbasic}
\bibliography{main}
}

\end{sloppypar}
\end{document}